\documentclass[10pt,letterpaper]{article}

\usepackage{cogsci}

\cogscifinalcopy
\usepackage[flushleft]{threeparttable}
\usepackage{amsmath}
\usepackage{tipa}

\usepackage{pslatex}
\usepackage{graphicx}
\usepackage{xcolor}
\usepackage{apacite}
\usepackage{float} 

\usepackage{CJKutf8}

\title{Around the world in 60 words: A generative vocabulary test for online research}

\author{
    {
    \large 
	Pol van Rijn\textsuperscript{1,}, 
	Yue Sun\textsuperscript{2,},
	Harin Lee\textsuperscript{1,3},
	Raja Marjieh\textsuperscript{4,}, 
	Ilia Sucholutsky\textsuperscript{4,},
    Francesca Lanzarini\textsuperscript{2,},
    } \\
    {
    \large
    \textbf{
	Elisabeth Andr{\'e}\textsuperscript{5,},
	and Nori Jacoby\textsuperscript{1}
    }
    }\\
   	\textsuperscript{1}Computational Auditory Perception Group, Max Planck Institute for Empirical Aesthetics, Frankfurt am Main, Germany \\
    \textsuperscript{2}Ernst Strüngmann Institute for Neuroscience, Frankfurt am Main, Germany \\
  	\textsuperscript{3}Max Planck Institute for Human Cognitive and Brain Sciences, Leipzig, Germany \\
   	\textsuperscript{4}Princeton University, New Jersey, USA\\
   	\textsuperscript{5}University of Augsburg, Augsburg, Germany\\
}

\begin{document}
\maketitle
\begin{abstract}

Conducting experiments with diverse participants in their native languages can uncover insights into culture, cognition, and language that may not be revealed otherwise. However, conducting these experiments online makes it difficult to validate self-reported language proficiency. Furthermore, existing proficiency tests are small and cover only a few languages. We present an automated pipeline to generate vocabulary tests using text from Wikipedia. Our pipeline samples rare nouns and creates pseudowords with the same low-level statistics. Six behavioral experiments (N=236) in six countries and eight languages show that (a) our test can distinguish between native speakers of closely related languages, (b) the test is reliable ($r=0.82$), and (c) performance strongly correlates with existing tests (LexTale) and self-reports. We further show that test accuracy is negatively correlated with the linguistic distance between the tested and the native language. Our test, available in eight languages, can easily be extended to other languages.
\end{abstract}

\section{Introduction}
Large-scale online experiments allow researchers to recruit diverse populations from around the world, thus reducing the ``WEIRD" (Western, Educated, Industrial, Rich, Democratic) recruiting bias in cognitive science \cite{barrett2020crosscultural, henrich2010weird}. However, in online experiments, there is often less control on recruiting compared with a lab study. Thus it is critical to quickly and accurately assess the language proficiency of participants and not, which is common research practice, blindly trust self-report questionnaires \cite{lemhoefer2011LexTALE}. 

Language proficiency is usually measured using vocabulary tests, which test the participants' knowledge of relatively uncommon words.
In contrast to the majority of vocabulary tests, which are designed in the context of second-language education \cite{alderson2005DIALANG}, the LexTALE test \cite{lemhoefer2011LexTALE} has been developed to meet the needs of behavioral lab experiments. In this short pen-and-paper test, participants are asked to go through the list and determine if each test item is a real word or a pseudoword (i.e. not a real word). Real words are infrequent words (occurring approximately 1--20$\times$ per million) which are assumed to be only recognized by highly proficient speakers. Pseudowords are created by changing letters in existing words or by recombining existing morphemes. In this test, participants should obtain a higher accuracy if they are more proficient in the language. This procedure makes the test faster than educational vocabulary tests, which can take up to 45 minutes to complete \cite{university2001qpt}.

 The test was originally developed for English, Dutch, and German, and was later adapted to Chinese \cite{chan2018LEXTALECH}, Italian \cite{amenta2020LexITA}, Finnish \cite{salmela2021Lexize}, French \cite{brysbaert2013LextaleFR}, Spanish \cite{izura2014LextaleEsp}, and Portuguese \cite{zhou2021LextPT}. Extending the test to another language however, requires extensive manual work and experimental validation.

While LexTALE was developed for laboratory experiments, it does not work well in online experiments. For example, LexTALE uses a fixed presentation order (participants who participate in multiple online experiments can memorize the order); it presents twice as many words as pseudowords (participants can exploit this imbalance); and items are not presented with a fixed pace (without time limitation participants can use an online dictionary search to look up the word). Here, we address these issues and additionally overcome two fundamental limitations in the LexTALE design. LexTALE relies on (1) the existence of curated word frequency databases and (2) the domain knowledge of the creator, which can be up to a certain degree of subjectivity (e.g., which linguistic rules were applied for the creation of pseudowords). These limitations have likely contributed to the fact that LexTALE is only available in a small set of languages and consists of a small number of words, which makes repeated participation impossible due to memory effects.

\begin{figure*}[ht!]
    \begin{center}
        \includegraphics{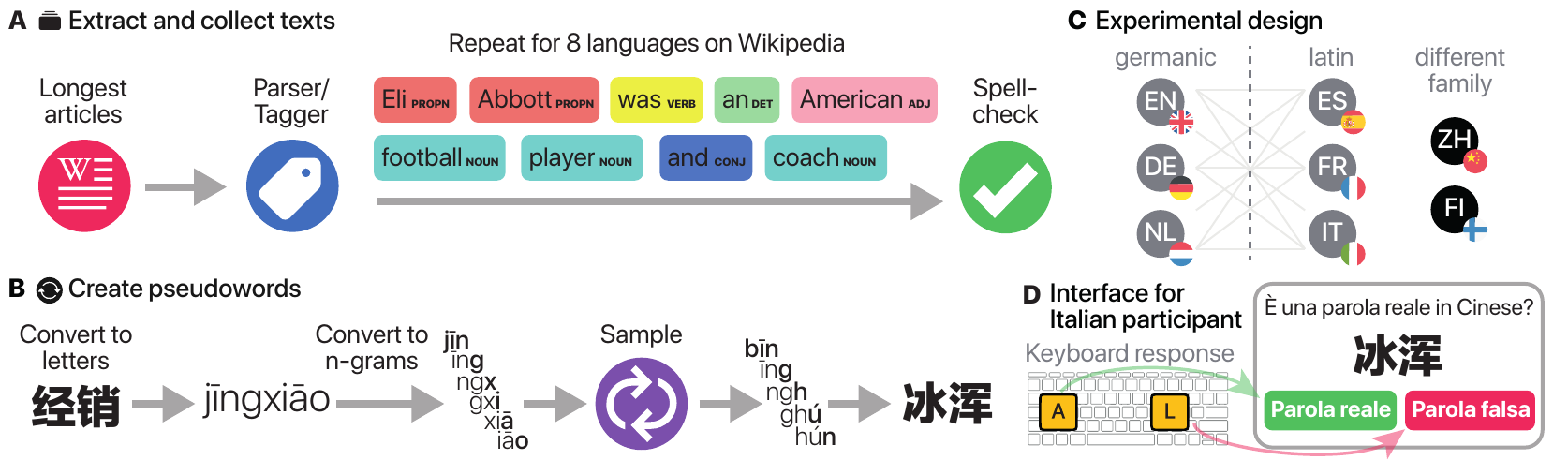}
    \end{center}
    \caption{Stimulus creation and study design. \textbf{A} Real words were extracted from text found on Wikipedia. Each token was lemmatized and spellchecked. \textbf{B} From all accepted lemmas in a language, we compute the n-gram transitional probabilities of letters within words. Pseudowords are created by drawing samples from the transitional probability. Here for demonstration purposes, we use 3-grams without * padding. Character-based languages, such as Chinese, are first converted into letter-based representations (e.g., Pinyin) to compute the transitional probabilities. The letters are later converted back to character sequences.  \textbf{C} About 40 monolingually-raised participants from the United Kingdom, Germany, the Netherlands, Spain, France, and Italy participate in a language test in their native (L1) and in a foreign language (L2). \textbf{D} Interface for an Italian participant. Participants use their keyboard for quick responses.}
    \label{fig:approach}
\end{figure*}

To alleviate these issues, we developed WikiVocab -- an automatic pipeline to generate vocabulary tests based on Wikipedia. Wikipedia is one of the largest collaborative internet projects and available in 329 languages\footnote{https://en.wikipedia.org/wiki/Wikipedia}. Because of this it, can be used to estimate vocabulary frequency in multiple languages. In our approach, we obtain a word frequency distribution from Wikipedia in a given language using automated heuristics. We sample rare words from this distribution, compute the n-gram transitional probabilities of letters within words, and create pseudowords that exhibit comparable n-gram probabilities. Since we obtain the word frequency distribution directly from a text corpus and implicitly capture orthographic regularities via n-gram probabilities, our approach does not rely on domain expertise nor on curated databases. The automatic pipeline allows us to create a very large pool of real words and pseudowords.

In what follows, we will describe how we generate the vocabulary test and benchmark it against available LexTALE tests in eight languages using six experiments with a total of 236 participants recruited from six countries. We show that our test can separate native speakers of closely related languages, that participants' performance in our test strongly correlates with LexTALE and self-reports, and finally, that participants' scores in different foreign languages reflect linguistic distances from these languages to their native language.

\section{Methods}
\subsection{Extract and collect texts}
\paragraph{Text processing} For each language, we process the longest 10,000 articles on Wikipedia (see Figure \ref{fig:approach}A), as the same articles in different languages fluctuate in their length. We use UDPipe 2.0 \cite{straka2018udpipe} to lemmatize all words and obtain a Part-Of-Speech tag (POS) to exclude proper nouns such as ``Eli Abbott" from the vocabulary test (see Figure \ref{fig:approach}A for the example). Since each Wikipedia page from a given language may contain words from multiple languages, we automatically check the original language of each word, using two approaches: (1) by using \texttt{fasttext} \cite{bojanowski2016fasttext} and (2) by using open-source spellcheckers from LibreOffice dictionaries\footnote{\url{https://github.com/LibreOffice/dictionaries}} using \texttt{guess\_language-spirit}\footnote{\url{https://pypi.org/project/guess\_language-spirit/}} and \texttt{pyenchant}\footnote{\url{https://github.com/pyenchant/pyenchant}}. A word is considered to be part of the associated language if either the spellchecker or \texttt{fasttext} correctly predicts that language. In addition to those filters, we add some language-agnostic preprocessing steps. For example, we reject words that are written in exclusively capital letters, exclude one-letter words, and remove words containing numbers or punctuation. From all words that fulfill these criteria, we only select the nouns to make the test more comparable across languages. In addition, we only use the lemmatized words to avoid multiple word forms of the same word.

\paragraph{Word filtering}
After we collect all the words per language, we convert all words to lowercase. Next, we remove words that exclusively occur in a small set of articles as these words tend to be jargon (e.g. `hippocampus') and are therefore not a good fit for a general-purpose vocabulary test. Thus, we calculate the ratio of word counts to the number of different articles in which the word occurs, and keep words in the 95th percentile, thus cutting off the top 5\%.

\paragraph{Compound word removal}
We train \texttt{charsplit} \cite{tuggener2016charsplit} on all the accepted lemmas in a language to detect potential compound words (for example ``snowball'' in English). We consider a word to be a compound word if there is a plausible word boundary between two words (i.e., $>$ 0; threshold proposed by \citeA{tuggener2016charsplit}) and if the last segment of the detected compound is a valid word in that language. The exclusion of compound words is relevant for many languages, as sampling from n-grams of compound words tends to generate even longer compound words, which are never used in the language, but do have a lexical meaning (for example ``Krankenversicherungsänderungsgesetzesentwurf" in German, which would mean: a draft of a law to change a health insurance provider). Another issue with compound words is that they pollute the word frequency distribution: compound words typically occur less often than each of their compounds, but this does not make the compound word harder to recognize than its elements (e.g., ``snowball" is not much harder than ``snow" and ``ball"). We therefore automatically remove all detected compound words. We could not apply compound word elimination to Chinese, as most multi-character words in Chinese would be considered compound words according to the heuristics above.

\subsection{Create pseudowords}
\paragraph{Convert to letters}
For Chinese, we use Pinyin\footnote{\url{https://github.com/mozillazg/python-pinyin}} to convert characters to sequences of letters from which we can obtain letters' n-grams (see Figure \ref{fig:approach}B). We store the mapping between characters and letters to convert the letters back to characters after sampling the n-grams.

\paragraph{Compute n-grams}
We compute n-grams of letter-based sequences from all words that are valid nouns and fulfill all filtering criteria. Existing linguistic work on spoken lexicons of multiple languages has shown that pseudowords generated from a 5-phone model capture most phonotactic regularities across the real words of the language \cite{dautriche2017_phonotactic_regularities, trott2020_phonotactic_regularities}. While, n-phone statistics on the phonetic transcriptions of words are not identical to n-gram statistics on written words\footnote{For example, a multi-letter grapheme $<$ph$>$ can refer to a single phoneme /f/, or certain graphemes are not pronounced such as the final $<$e$>$ in ``lak\textbf{e}".}, n-phones are similar to n-gram statistics. We, therefore, use 5-gram transitional probabilities to create pseudowords, since they are the closest equivalents to 5-phone transitional probabilities for the written language. To be able to create pseudowords that contain less than 5 letters, we pad each word with four asterisks on each side.

\paragraph{Sample from n-grams}
We start by sampling a 5-gram sequence which starts with four asterisks (`****') as the start symbols of a word. When sampling the next 5-gram, we make sure that the first four letters are identical to the last four letters of the previous 5-gram. We repeat this procedure until a 5-gram is drawn, which ends with an asterisk that indicates the end of the word. We remove the asterisks and validate the created string of letters. We keep on sampling until we create 1,000 unique pseudowords.

\paragraph{Validate pseudowords}
We reject generated pseudowords that correspond to real words in the language. We also reject pseudowords that contain too few or too many letters based on the range of word length of all real words of each language. For Chinese, we convert each pseudoword from the Pinyin representation back to the character representation. We do this by replacing all characters. To replace the longest letter sequences first, we sort the letter-character mapping by the length of the letter string. If there are left-over letters in the pseudowords, the word is rejected. For all other languages, we check if the created pseudowords are likely to be compound words. We reject the word if this is the case. To avoid the creation of pseudowords that look similar to existing words and are potential typos, we compute a fuzzy search using \texttt{thefuzz}\footnote{\url{https://github.com/seatgeek/thefuzz}}. Since the total number of lemmas and tokens is extremely large, we limit the search to words that start with the first and last three letters and are of a similar length (10\% difference in length allowed). We store the maximum match between the pseudowords and any word in that language.

\paragraph{Identify difficult words}
To match the task difficulty of LexTALE, we first look up the real LexTALE items in our word frequency distribution (log10-scale). We then  compute the mean and standard deviation of the LexTALE items per language. We select real words by sampling from a normal distribution centered at the average LexTALE word frequency with the standard deviation being computed over all languages with LexTALE.

\paragraph{Pair words}
Out of the 1,000 created pseudowords, we select 500 that best match the words in that language. To do so, we obtain the logarithm of the transitional probabilities of the letters' 5-grams for both the words and the pseudowords. We then computed the average absolute difference between the words and pseudowords that have the same number of 5-grams. On the resulting distances, we keep matching the word and pseudowords with the smallest distance. So, we match words and pseudowords that have a similar `rarity' at the same positions in the word. We repeat this procedure until we match all pseudowords. From the matched list, we only include 500 matched pseudowords that have the smallest fuzzy match ratio to any of the words in that language. By doing so, we select the 500 pseudowords which are least likely to be typos. For compatibility with LexTALE, we only include the first 60 items for the behavioral experiment.

\subsection{Experimental Design}
We selected eight languages to benchmark our test on (see Figure \ref{fig:approach}C). Six languages are Indo-European languages and can further be grouped into Germanic (English, \texttt{en}; German, \texttt{de}; Dutch, \texttt{nl}) and Italic (Spanish, \texttt{es}; French, \texttt{fr}; Italian, \texttt{it}) subfamilies. Chinese (\texttt{zh}) is a Sino-Tibetan language and Finnish (\texttt{fi}) is a Uralic language. We use a semi-balanced design, in which native speakers of the six Indo-European languages participate in the experiment. Each participant perform LexTALE and WikiVocab in their native language and in one of the other seven foreign languages selected at random. The order of the tests is also randomized. One real word and pseudoword for each language are presented in Table \ref{tab:example_items} as examples.\footnote{The list of all the stimuli can be downloaded here: \url{https://cogsci-23.s3.amazonaws.com/wikivocab.zip}.}

\begin{table}[ht]
\centering
\begin{tabular}{rll}
  \hline
 & real & pseudo \\ 
  \hline
  de &  fasanerie & fälschuld  \\ 
  nl & verrechtsing & rugzeeschaf \\ 
  en & carrion & washioneer  \\ 
  fr & envasement & paléonton \\ 
  it & sciocchezza & giustibola  \\ 
  es & demonio & tonomía \\ 
  fi & figuuri & antalaisu  \\ 
  zh & \begin{CJK*}{UTF8}{gbsn}经销\end{CJK*}  &\begin{CJK*}{UTF8}{gbsn}冰浑\end{CJK*}  \\ 
   \hline
\end{tabular}
\caption{Example items}
\label{tab:example_items}
\end{table}

\subsection{Participants}
To collect the data in the six countries, we ran online experiments on Prolific\footnote{\url{https://www.prolific.co}}. All texts in the interface of the experiment (e.g., buttons, instructions, etc) were presented in the native language of the participant. Non-English texts were first automatically translated using DeepL, then manually checked and corrected by a native speaker of each language. The recruitment and experimental pipelines were automated using PsyNet~\cite{harrison2020psynet}\footnote{\url{https://www.psynet.dev}}, a modern framework for experiment design and deployment which builds on the Dallinger\footnote{\url{https://dallinger.readthedocs.io/en/latest/}} platform for recruitment automation. Participants have to be born and currently reside in the targeted country, raised monolingually, hold nationality from that country, and speak the target language as their mother tongue. Overall, 236 participants with a mean age of 33 (SD = 12) complete the study. 61 \% of the participants have at least a Bachelor's degree. Participants were paid 9 GBP per hour and provided informed consent according to an approved protocol.

\subsection{Interface}
Words and pseudowords were presented in random order. To reduce the chance that a participant will search the word on the internet, we disabled the ability to copy the target word and limited the display time to two seconds. Participants were asked to respond as fast as possible by pressing two dedicated keys on their keyboard (see Figure \ref{fig:approach}D). To estimate the reliability of the test, participants did two batches of trials per language and test. Each batch contained 30 trials.

\begin{figure*}[ht!]
    \begin{center}
        \includegraphics{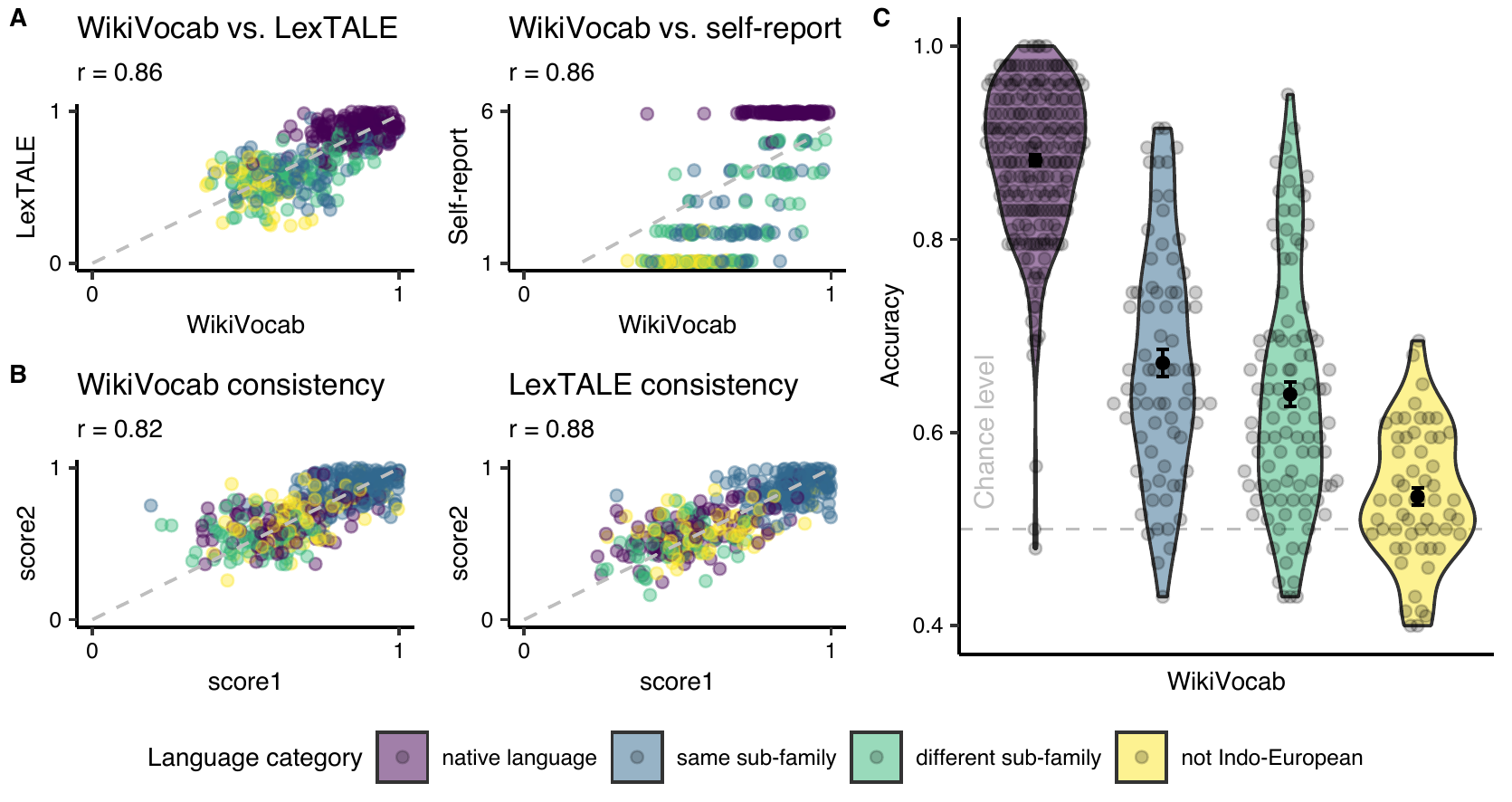}
    \end{center}
    \caption{Results. \textbf{A} Scatter plots of WikiVocab accuracy compared with LexTALE accuracy (left) and language self-report (right). \textbf{B} Consistency in the accuracy of the same participant for WikiVocab (left) and LexTALE (right). \textbf{C} WikiVocab accuracy on native language (i.e., L1), same language sub-family (e.g., L1 = Italian, L2 = Spanish), different sub-family (e.g., L1 = English, L2 = Spanish), and not Indo-European (e.g., L1 = German, L2 = Finnish). The dots represent measurements from all languages. All error bars are standard errors from the mean.}
    \label{fig:results}
\end{figure*}

\section{Results}
We computed the correlation between the average performance in the two blocks for WikiVocab and LexTALE for each language per participant. As shown in Figure \ref{fig:results}A the performance in LexTALE strongly correlates ($r = 0.86$, $p < .001$) with the performance in WikiVocab. The score in WikiVocab also strongly correlates with the language self-report of the participant ($r = 0.86$, $p < .001$). To measure the consistency on the test, we measured the correlation between performances on both blocks: $r = 0.82$ ($p < .001$) for WikiVocab (Figure \ref{fig:results}B) which is similarly high as the consistency measured for LexTALE ($r = 0.88$,  $p < .001$). In Figure \ref{fig:results}D, we show that the accuracy of the native language is much higher than for the other foreign languages.

To further benchmark the human performances on WikiVocab, we also ran the same test on GPT-3 (see Table \ref{tab:accuracies}), a large language model pre-trained on a massive corpus of language data~\cite{brown2020language}. To elicit GPT-3 responses we used the following prompt ``\textit{You will be shown a series of words and must determine if each one is a real or fake word in \{language\}. If it is a real word in \{language\} respond with 1. Otherwise, respond with 0. 
Word: \{word\}
Response:}", where \{word\} was replaced by the target word and \{language\} by the target language.
GPT-3 outperformed native speakers in all languages except for Dutch and German. Both languages unite a strong tendency to create long compound words. In Dutch, 9/14 mistakes were made because, by misclassifying, the residual compound words were not detected as real words. Importantly, GPT3 performed nearly perfectly on what can be seen as its ``L1" (accuracy in English was 0.98) compared with lower performances on its ``L2s"  (performance in other languages ranges between 0.67-0.95). Moreover, languages that are likely to be well-represented in the training set of GPT-3 (es, fr and it) have relatively high scores (larger than 0.88) suggesting that ``language familiarity" also affects task performance for GPT-3.

\begin{table}[ht]
\centering
    \begin{tabular}{rlrrrrrr}
      \hline
    & DE & NL & GB & FR & IT & ES & GPT-3 \\ 
      \hline
      de & \textbf{0.90} & 0.69 & 0.52 & 0.53 & 0.60 & 0.53 & 0.85 \\ 
      nl & 0.55 & \textbf{0.84} & 0.65 & 0.54 & 0.54 & 0.56 & 0.77 \\ 
      en & 0.82 & 0.84 & \textbf{0.86} & 0.84 & 0.84 & 0.79 & 0.98 \\ 
      fr & 0.59 & 0.65 & 0.58 & \textbf{0.83} & 0.66 & 0.60 & 0.88 \\ 
      it & 0.61 & 0.51 & 0.60 & 0.66 & \textbf{0.92} & 0.70 & 0.92 \\ 
      es & 0.72 & 0.65 & 0.74 & 0.69 & 0.73 & \textbf{0.93} & 0.95 \\ 
      fi & NA & 0.51 & 0.51 & 0.53 & 0.50 & 0.45 & 0.73 \\ 
      zh & 0.57 & 0.54 & 0.55 & 0.57 & 0.56 & 0.52 & 0.67 \\ 
       \hline
    \end{tabular}
    \caption{Human vs. GPT-3 performance on WikiVocab. The rows are the tested languages, the columns are the country of the native speakers. Note: No German participant completed the Finnish WikiVocab task, so there is a missing data point.}
    \label{tab:accuracies}
\end{table}

\begin{figure*}[ht!]
    \begin{center}
        \includegraphics{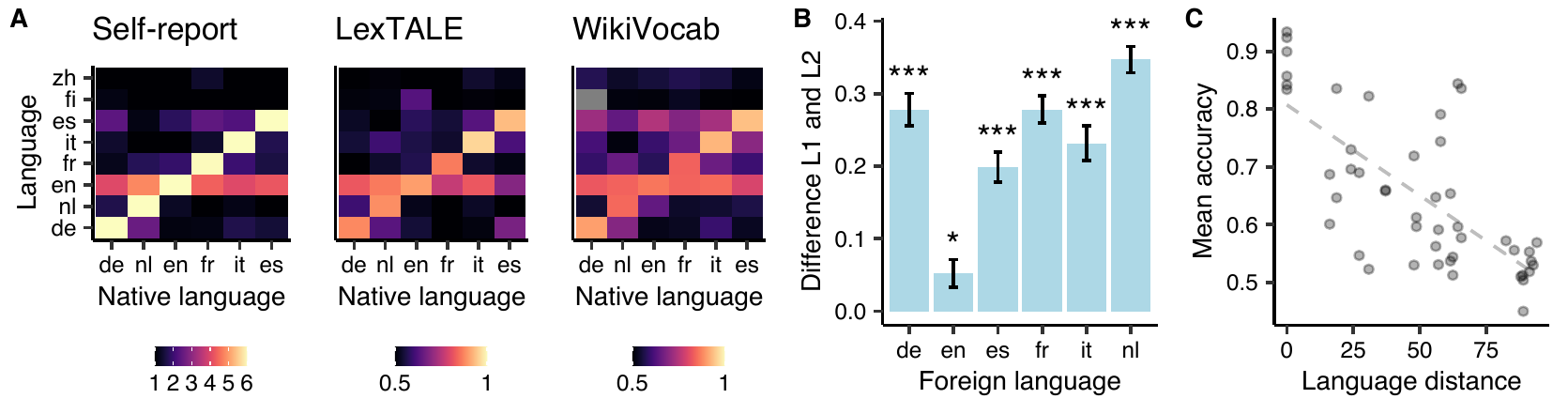}
    \end{center}
    \caption{Results across languages. \textbf{A} Heatmaps of the mean self-report (left), LexTALE accuracy (center), and WikiVocab accuracy (right). \textbf{B} For each of the native languages (x-axis), we computed the difference in performance compared to the tested foreign language. The asterisks indicate the level of significance of a paired t-test. Significance codes: * = $p < 0.05$, *** = $p < 0.001$. Error bars represent the standard error of the mean. \textbf{C} Scatter plot of linguistic distance between the tested and native language and mean WikiVocab accuracy.}
    \label{fig:results_across}
\end{figure*}

In Figure \ref{fig:results_across}A, we show the average self-report proficiency and the performance on LexTALE and WikiVocab in the foreign and native languages. Interestingly, most participants score well on the English test and also indicate in the self-report they are proficient in English. Other than that, in all three cases, the diagonals (reflecting L1) are higher than non-diagonal items.

To assess whether participants perform better in their native language compared with a foreign language, we computed the difference between their performances (Figure \ref{fig:results_across}B). We found  that the gap between one's native language and the foreign language is largest for Dutch, whereas the difference in performance between native languages and English was lowest. The strong horizontal line in Figure \ref{fig:results_across}A for English shows that most participants in the test were fluent in English and the gap to L1 (the diagonal) was smallest. Nonetheless, for all languages, the performance in the native language was significantly higher than in the foreign language ($p<0.05$).

Finally, we correlated the mean accuracy on the test with linguistic distance between the tested and native language from \citeA{beaufils2020lexicaldistance}. Test score and linguistic distance are negatively correlated $r = -0.71$ ($p < 0.001$), indicating that the performance on the test is lower if the tested language is further away from one's native language (so German and Dutch are ``closer'' in linguistic distance compared with Dutch and Finnish).

\section{Discussion}
In this paper, we propose a new approach, ``WikiVocab'', to create automated vocabulary tests for online research. We generated the test for eight languages and showed that our test allows us to dissect between native speakers of closely related languages and that the performance strongly correlates with existing tests and self-reports. When looking into the differences between the languages, we showed that the test performances correlate with linguistic distance and that the native language performance is significantly higher than the same participants' performance in foreign languages.

Our results show that most participants are highly proficient in English, in their self-reports and on both tests. One possible explanation for this behavior might be the relatively high degree of formal education of the participants ($>$ 60\% have at least a Bachelor's degree) and the fact that most experiments on Prolific are in English. To increase the contrast between native and second language learners of English, one could further decrease the frequency of occurrence of real words which makes them more difficult to recognize.

The pairwise comparisons between the countries also reveal that the findings are strongly directional. Whereas the Dutch participants almost reach 70\% accuracy in the German test, German participants are at chance-level for the Dutch tests. These differences might be attributed to which languages are taught in school in these two countries. Generally, we can expect an increased accuracy for languages that are part of the curriculum in the respective countries. 

GPT-3 generally performs well on most languages, indicating that large language models can outperform humans on the task and bots can potentially mimic human behavior. One way to avoid this would be to present text as images, continue preventing copying, limit viewing and response times, and possibly add modifications similar to CAPTCHA \cite{bursztein2010captcha}. 

One important limitation of our method is it relied on linguistic resources (such as parsers, lemmatizers, and spell checkers). We found that these tools are available in at least 60 languages, suggesting that our method may apply to a large number of languages. For low resources languages where these are not available, in future work we plan to test if removing some of the processing steps would still allow good task performance. Alternatively, it is possible to recruit native speakers to test performance on candidate words and select only words that can be reliably detected by native speakers.

To conclude, WikiVocab offers a systematic and data-driven way to automatically produce language tests, which can be used to screen participants, particularly in the online environment. The generative nature of the test allows researchers to increase the diversity of  online research and we believe would be an important tool to assess language proficiency in human behavioral experiments. More broadly, our work shows how to scale experimental methods from the lab to the online world, thus contributing to the scale and diversity of cognitive science.

\bibliographystyle{apacite}

\setlength{\bibleftmargin}{.125in}
\setlength{\bibindent}{-\bibleftmargin}

\bibliography{main}

\end{document}